\documentclass[letterpaper, 10 pt, conference]{ieeeconf}  

\IEEEoverridecommandlockouts                              

\usepackage{xcolor}

\usepackage{graphicx}

\usepackage{gensymb}
\usepackage[hyphens]{url}
\usepackage{multirow}
\usepackage{multicol}
\usepackage{tabularx}
\usepackage{dsfont}
\usepackage{url}
\usepackage{color}
\usepackage{subcaption}
\usepackage{caption}
\usepackage{booktabs}
\usepackage{dcolumn}
\usepackage{amssymb}
\usepackage{hyperref}

\hypersetup{
    colorlinks=false,
    linkcolor=blue,
    filecolor=magenta,      
    urlcolor=cyan,
    pdftitle={Overleaf Example},
    pdfpagemode=FullScreen,
    }

\title{GIDP: Learning a Good Initialization and Inducing Descriptor Post-enhancing for Large-scale Place Recognition}

 \author{Zhaoxin Fan$^{1}$, Zhenbo Song$^{2}$, Hongyan Liu$^{3}$, Jun He$^{1}$ 
 \thanks{$^{1}$Renmin University of China, {\{\tt\small{fanzhaoxin, hejun}\}}{\tt\small{@ruc.edu.cn}}}
 \thanks{$^{2}$Nanjing University of Science and Technology, {\{\tt\small{songzb}\}}{\tt\small{@njust.edu.cn}}} 
 \thanks{$^{3}$Tsinghua University, {\{\tt\small{hyliu}\}}{\tt\small{@tsinghua.edu.cn}}} 
}

\begin{document}

 \maketitle
\thispagestyle{empty}
\pagestyle{empty}

\begin{abstract}
Large-scale place recognition is a fundamental but challenging task, which plays an increasingly important role in autonomous driving and robotics. Existing methods have achieved acceptable good performance, however, most of them are concentrating on designing elaborate global descriptor learning network structures. The importance of feature generalization and descriptor post-enhancing has long been neglected. In this work, we propose a novel method named \textbf{GIDP} to learn a \textbf{G}ood \textbf{I}nitialization and Inducing \textbf{D}escriptor \textbf{P}ose-enhancing for Large-scale Place Recognition. In particular, an unsupervised  momentum contrast point cloud pretraining module and a reranking-based descriptor post-enhancing module are proposed respectively in GIDP. The former aims at learning a good initialization for the point cloud encoding network before training the place recognition model, while the later aims at post-enhancing the predicted global descriptor through reranking at inference time. Extensive experiments on both indoor and outdoor datasets demonstrate that our method can achieve state-of-the-art performance using simple and general point cloud encoding backbones.

\end{abstract}

\section{INTRODUCTION}
Large-scale place recognition is a challenging but important task, which plays an increasingly important role in autonomous driving \cite{xu2022v2x,xu2022opv2v,xu2022cobevt} and robotics \cite{alatise2020review,belanche2020service,huang2021engaged}.  The place recognition result is always used to provide the agent with accurate localization information. For instance, suppose there is a robot waiter in a hotel who is asked to transport luggage to the guest's room. To locate the robot itself,  place recognition is always required, especially in an indoor environment where GPS signal is not available.

For large-scale place recognition task,  a HD map of the zone where the robot serves is always pre-built. Then, the HD-map is cut into thousands of different submaps. The localization information of each submap in the HD map is known information. When the robot walks in the zone, it compares the current captured scene with submaps in the database and finds the most similar one. To this end, the location of current scene is equivalent to the known location of the most similar submap.
\begin{figure}[!t]
\centering
\includegraphics[width=\linewidth]{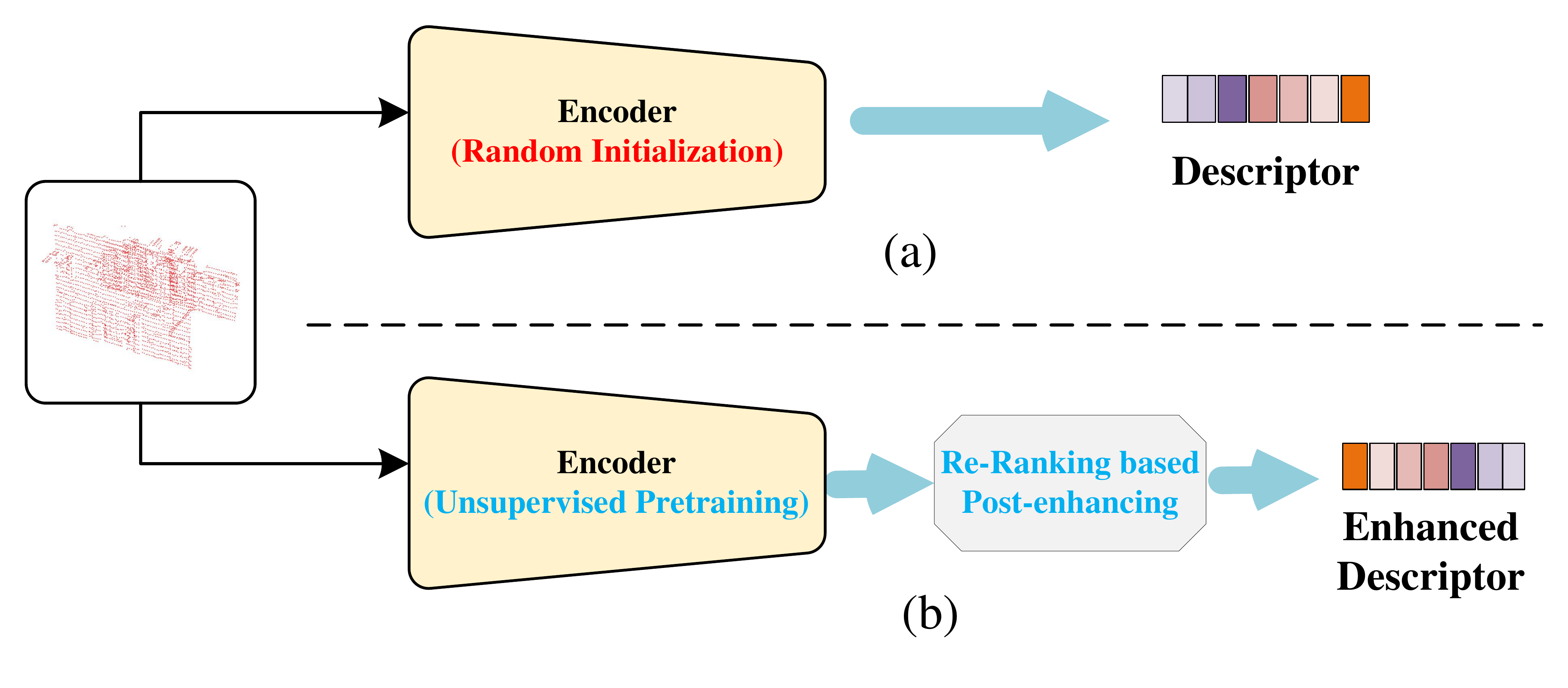}
\caption{The different between our work and previous methods. (a) Training and inference pipeline of previous methods. (b) Training and inference pipeline of our GIDP.}
\label{fig:intuition}
\end{figure}
A straight forward way to achieve the above place recognition task is to use images to represent scenes \cite{arandjelovic2016netvlad,yu2019spatial,hausler2021patch}. However, images are sensitive to environmental changes such as weather change and season change \cite{fan2022svt}. Compared to image, point cloud captured by LiDAR is much more robust towards variants caused by the above environmental changes.  Therefore, in this paper, we study the more robust point cloud based large-scale place recognition.

PointNetVLAD \cite{uy2018pointnetvlad} is the first deep learning based work for large-scale place recognition. It uses PointNet \cite{qi2017pointnet} and NetVLAD \cite{arandjelovic2016netvlad} to learn point cloud descriptors. Then, following works \cite{zhang2019pcan, sun2020dagc,xia2021soe,liu2019lpd,fan2020srnet, komorowski2021minkloc3d,fan2022svt} try to improve the performance through the perspective of using graph networks \cite{sun2020dagc,fan2020srnet,liu2019lpd}, designing attention mechanism \cite{zhang2019pcan,xia2021soe}, and using sparse convolution \cite{komorowski2021minkloc3d,fan2022svt}. Though all of them achieve remarkable performance, the above mentioned methods all focus on designing novel network architectures for better global point cloud descriptor learning. The equally important network weights initialization and descriptor post-processing have long been neglected as shown in Fig. \ref{fig:intuition} (a).

To tackle the above issue, we propose a novel method named \textbf{GIDP} to learn a \textbf{G}ood \textbf{I}nitialization and Inducing \textbf{D}escriptor \textbf{P}ose-enhancing for Large-scale Place Recognition as shown Fig. \ref{fig:intuition} (b). The intuition behind GIDP is that we support the opinion that learn a pretraining model tailored for point cloud-based large-scale recognition is very important. Moreover, we are also motivated by the fact that the learned global descriptors can be further enhanced before similarity calculation. To this end, GIDP consists of two main modules. The first one is an unsupervised  momentum contrast point cloud pretraining module, which aims at learning a good initialization for the descriptor encoding network. Benefited from the power of our pretraining module, a simple point cloud encoding network such as PointNet \cite{qi2017pointnet} and DGCNN   \cite{wang2019dynamic} can achieve promising result. Result in a more light-weight network architecture and more efficient inference than existing methods. The second module is a re-ranking based descriptor post-enhancing module. In this module, a descriptor is interacted with other descriptors in an simple yet effective manner and then re-ranked at inference time. In this way, the descriptor adopts more semantic information to  describe the scene  and finally result in performance improvement. To the best of knowledge, we are the first to design pretraining models and post-processing modules for point cloud-based large-scale place recognition.

We conduct extensive experiments on several widely used indoor and outdoor datasets. Results show that our method achieves new state-of-the-art performance using simple backbones. Our contributions can be summarized as follows:
\begin{itemize}
    \item We propose GIDP, a novel point cloud based large-scale place recognition method, which achieve state-of-the-art performance using simple backbones. 
    \item We propose  an unsupervised  momentum contrast point cloud pretraining module and a re-ranking based descriptor post-enhancing module to improve the place recognition performance.
    \item We conduct extensive experiments on both indoor and outdoor datasets to show the effectiveness of our method and the proposed modules.
\end{itemize}

\section{Related Work}

In our work, in this section, we introduce the related work. Since we aim at improving the performance of point cloud-based large-scale place recognition methods, we first introduce the state-of-the-art place recognition models. Then, we present recently popular unsupervised training methods. Finally, we introduce related re-ranking methods.

\subsection{Large-scale place recognition} 

Large-scale place recognition can be divided into image-based methods \cite{arandjelovic2016netvlad,yu2019spatial,hausler2021patch} and point cloud based methods \cite{zhang2019pcan, sun2020dagc,xia2021soe,liu2019lpd,fan2020srnet, komorowski2021minkloc3d,fan2022svt}. Image based methods are proven to be sensitive to environmental change, hence attract less attention in recent year. In contrast, point cloud-based methods are very robust benefited from the inherent properties of LiDAR. PointNetVLAD \cite{uy2018pointnetvlad} is the first deep learning method for point cloud large-scale place recognition simply uses PointNet \cite{qi2017pointnet} and NetVLAD \cite{arandjelovic2016netvlad}. Then, DAGC \cite{sun2020dagc} and SR-Net \cite{fan2020srnet} utilizes dynamic graph network and static graph network respectively to for this task, while SOE-Net \cite{xia2021soe} and PCAN \cite{zhang2019pcan} propose different attention mechanism for this task. LPD-Net \cite{liu2019lpd} uses hand-craft features to further improve performance. Recently, sparse convolution \cite{choy20194d} becomes popular, therefore, methods like \cite{komorowski2021minkloc3d,fan2022svt} are proposed to use sparse convolution to learn point cloud descriptors. Though promising, all the above proposed methods are trying to design better network structures for performance improvement. In contrast, we propose to learn a good initialization and good post-enhancing to improve place recognition performance.




\subsection{Unsupervised pretraining} 
Unsupervised pretraining has become very popular in recent years due to its effectiveness. In the natural language processing (NLP) field,  BeRT \cite{devlin2018bert} is a phenomenal work that utilizes unsupervised pretrain to train a transformer model, which now is the most widely used NLP backbone. In computer vision field, MOCO V1/V2 \cite{he2020momentum, chen2020improved} propose to use contrastive learning for pretraining and use data augmentation to generate positive and negative samples. SimCLR \cite{chen2020simple} researches the combination of different data augmentation methods and proposes a projection head in its pretraining framework. DINO \cite{caron2021emerging} introduces an unsupervised pretraining framework and finds that the attention map can well describe the foreground objects. The above mentioned methods are all image-level methods, while DenseCL \cite{wang2021dense} presents a dense contrastive training pipeline to densely pretrain the network in pixel level, which is more suitable for dense tasks such as segmentation. In point cloud processing, there are also some works \cite{zhang2021self,xie2020pointcontrast} are proposed using contrastive learning. In our work, we also use contrastive learning for unsupervised point cloud encoding network pretraining. Our method is tailored for large-scale place recognition.

\subsection{Re-ranking methods} 
In the current large-scale place recognition framework, the task is essentially a retrieval problem. In retrieval, re-ranking of learned descriptors is a hot topic. \cite{zhong2017re} proposes K-reciprocal feature, which encodes the K-
reciprocal features into a singe vector for re-ranking. \cite{wang2019enhancing}  proposes a CNN semantic re-ranking system that greatly improves the retrieval performance. \cite{shen2021re} introduces a method to meta-learn the re-ranking updates for image retrieval. In UED \cite{bai2019re}, the re-ranking step is involved by running a diffusion process on the underlying data manifolds.  In the filed of NLP, Rocketqav2 \cite{ren2021rocketqav2} proposes to adaptively improve the  retriever and the re-ranker  according to each other’s relevance information. Though the above proposed methods demonstrate good performance, they are designed for images or texts. In our work, we propose a simple yet effective method for re-ranking descriptors learned from point clouds for the large-scale place recognition task.

\section{Methodology}
\subsection{Problem Formulation}
Given a HD-Map of a zone, we firstly cut it into a database of submaps: $\mathcal{D}=\{M_1, M_2, \cdot\cdot\cdot, M_n\}$, where $M_i$ is the $i-th$ submap in database $\mathcal{D}$.  Note the localization information of $M_i$ is known information. Then, assume that a robot is walk in the zone. The robot is asked to capture a scene $S \in R^{N \times 3}$ represented as point cloud at each walking step. Our goal is to find the most similar submap $M_s$ of $S$. Then, the location of $S$ is equal to the location of $M_s$. To do so, we learn a deep learning model $\mathcal{F}$ to encode $S$ and $M_i$ to a vector $v_s$ and $v_i$. Then, a KNN algorithm $\phi$ can be used to find $M_s$. The whole process can be defined as:
\begin{equation}
M_s=\phi(\mathcal{F}(S),\mathcal{F}(M_1, M_2, \cdot\cdot\cdot, M_n))
\end{equation}

\begin{figure*}[!t]
\centering
\includegraphics[width=\linewidth]{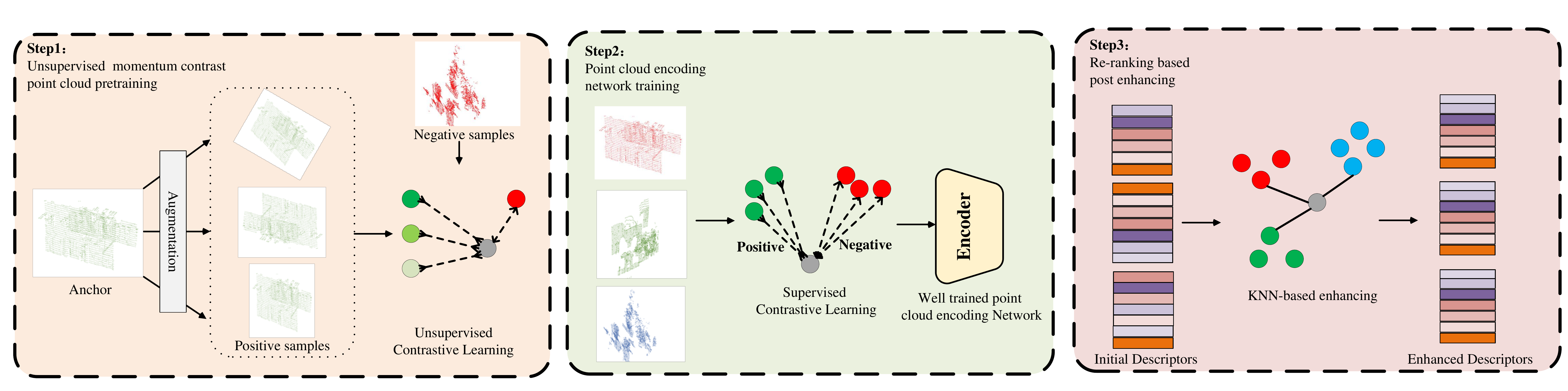}
\caption{Pipeline of our method. We use a simple backbone along with a GeM pooling module to consist of the point cloud encoding network. To train the network, an unsupervised momentum contrast point cloud pretraining module is proposed to first pretrain the network. Then, we supervisely train the network using contrastive learning. Finally,  we introduce a re-ranking based descriptor post-enhancing module to improve the powerful of predicted global descriptors.}
\label{fig:pipeline}
\end{figure*}

\subsection{Overview}
As mentioned before, training a point cloud encoding network and post-processing the predicted global point cloud descriptors are the most important factors of obtaining an excellent large-scale place recognition model. Therefore, in this work, we mainly research the two factors.

Fig. \ref{fig:pipeline} illustrates the pipeline of our work GIDP. As shown in the figure, our work can be divided into three stages. The first stage is an unsupervised  momentum contrast point cloud pretraining module. This module takes a batch of point clouds as input and uses contrastive learning to unsupervisely pretrain the point cloud encoding network. Then, in the second stage, we use the weights pretrained in the last stage to initialize the network and construct a supervised contrastive learning framework to train the network. In this stage, a GeM pooling \cite{} is used to aggregate global descriptor. After training, we can use the trained network to predict point cloud descriptor for place recognition. However, we find  the output descriptor is not powerful enough for large-scale place recognition. Therefore, at inference time, we use the third stage, the re-ranked based post-enhancing module to enhance the representational ability of the descriptor. Finally, a K-Nearest-Neighbor (KNN) algorithm can be used to find the most similar submap for each scene using these enhanced (or called re-ranked) descriptors. Next, we introduce the detail of the unsupervised  momentum contrast point cloud pretraining module, the re-ranking based post enhancing module and the loss function used to train the network.

\subsection{Unsupervised  momentum contrast point cloud pretraining module}
Though existing large-scale place recognition methods have achieved promising performance, there is still much room for their performance improvement. The main reason is that existing methods training the point cloud encoding network from scratch, which is easy to cause over-fitting. The consequence is that the abundant scene semantic and scene geometry  hidden in the point cloud can not be fully parsed, and so the finally predicted descriptors will be less effective. A possible solution is to pretrain the network before training for the particular place recognition task. Hence, the network would learn a good initialization where the feature is more powerful and more general. In this section, we introduce such a pretraining module named unsupervised  momentum contrast point cloud pretraining module to tackle the issue.

\noindent \textbf{Constructing training triplets:} Utilizing contrastive learning for pretraining is proven to be very effective. For every anchor point cloud $P_a$, at least a positive anchor $P_{pos}$ and a negative anchor $P_{neg}$ are required. However, since we aim at pretraining the network unsupervisely, there is no ground-truth information we can use except the point cloud itself to construct the \{anchor, positive sample, negative sample\} triplet. To this end, we propose to construct the triplet by data augmentation as introduced in \cite{chen2020simple}. Specifically, for $P_a$, we apply random data augmentation to generate its positive samples. Then, any other point clouds in the dataset can be its negative samples. Since we design the pretraining module for the large-scale place recognition task, the data augmentation should be carefully designed to maintain the scene semantic and geometry. Therefore, the specific data augmentation we choose includes: random jitter, random points removal, random block removal, and random shear.

\noindent \textbf{Contrastive learning:} Taking a batch of point cloud as input, suppose the output of the point cloud encoding network is $\{v_a, v_{p,1}, v_{p_2},\cdot\cdot\cdot, v_{p,m_p}, v_{n,1}, v_{n,2}, \cdot\cdot\cdot, v_{p,m_n}\}$, where $v_a$ is the descriptor of an anchor point cloud, $v_{p,i}$ is the descriptor of the positive sample and $v_{n,i}$ is the descriptor of the negative sample. We follow \cite{chen2020simple} to use a projection head to project $v_i$ to $u_i$,i.e., $\{u_a, u_{p,1}, u_{p21},\cdot\cdot\cdot, u_{p,m_p}, u_{n,1}, u_{n,2}, \cdot\cdot\cdot, u_{p,m_n}\}$. The projection head is implemented as a MLP layer. Then, for $v_a$, we choose one positive sample $u_{p,+}$ and $K$ negative samples to calculate the InfoNCE \cite{oord2018representation} loss:
\begin{equation}
L_{pretrain}=-log \frac{exp(u_a \cdot u_{p,+})}{\sum_{1}^{K}exp(u_a \cdot u_{n,k})}
\end{equation}

\noindent \textbf{Momentum update:} Following MOCO \cite{he2020momentum}, we record all descriptors as  a queue to calculate the InfoNCE  loss, which can make the dictionary large, therefore the back-propagation will not be limited by the information in a mini-batch. However,  it also makes it intractable to update the point cloud encoding network during back-propagation. To this end, we also use the momentum
update to address this issue. Specifically, we use two point cloud encoding networks during training, one is the anchor encoder which takes the anchor point cloud as input, and the another one is the momentum encoder which takes the positive samples and negative samples as input. The two encoders share the same network structure. Suppose the parameters of the anchor encoder is $\theta_a$ and the parameters of the momentum encoder is $\theta_{pn}$. We update  $\theta_a$ by back-propagation while update $\theta_{pn}$ by:
\begin{equation}
\theta_{pn} \leftarrow m \theta_{pn}+(1-m) \theta_a
\end{equation}

Without bells and whistles, through the above process, the point cloud encoding network can be comprehensively pretrained. Hence, performance of the downstream large-scale place recognition task can be greatly improved, evidenced by experimental results.

\subsection{Re-ranking based descriptor post-enhancing module}
After the unsupervised pretraining stage and the supervised encoding network training stage, the point cloud encoding model owns the ability of predicting discriminative global point cloud descriptors. However, the information hidden in the training dataset is still fully utilized. That is to say that the descriptor can still be enhanced using the training dataset. Motivated by this, in this section, we introduce a re-ranking based descriptor post-enhancing module, which can utilize the training set to further improve the power of descriptors.

\noindent \textbf{Related descriptors exploration:} For a scene presented by point cloud $P_a$, suppose its descriptor is $v_a \in R^C$, we argue that other descriptors $v_{o,i} \in \mathcal{D}$ in the training set share some general semantic and geometric features with it. Some of these features can be used to enhance $v_a$. To utilize such kind of information, we should first find the $K$ most relevant descriptors of  $v_a$. In our work, we use a KNN algorithm in the latent high-dimensional feature space to find them:
\begin{equation}
\{v_{o,1}, v_{o,2}, \cdot\cdot\cdot, v_{o,k}\}=KNN(v_a,\mathcal{D})
\end{equation}

\noindent \textbf{Inverse distance based enhancing:} After finding $\{v_{o,1}, v_{o,2}, \cdot\cdot\cdot, v_{o,k}\}$, we need to use them to enhance $v_a$. Note since we use contrastive learning to train the encoding network, theoretically, similar descriptors would locate close in the feature space while dissimilar ones are far away from each other. Therefore, distance in the feature space is an important factor to reflect the descriptor relationship. To this end, we propose an inverse distance based enhancing method to utilize the distance relationship of descriptors to enhance $v_a$. The enhancing process can be formulated as:
\begin{equation}
\hat{v_a}=\lambda v_a+ (1-\lambda){\sum_{1}^{K}( w_k \cdot v_{o,k})}
\end{equation}
where $w_k=\frac{exp(-|v_a-v_{o,k}|)}{\sum_{1}^{K}exp(-|v_a-v_{o,i}|)}$,  $\hat{v_a}$ is the enhanced global descriptor, and $\lambda$ is a balance term.

\noindent \textbf{Inductive vs Transductive:} Since the training set is pre-collected, it is very straight forward to re-use it at inference time for enhancing the descriptor in an on-line manner. This is in essence an inductive setting. We also note that in some cases, at inference time, we will collect many scenes at different location and then  retrieve the the location of each scene in an off-line manner. In such case, other scene collected at inference time can also be additionally used to enhance the descriptor. This inference time enhancing is called a Transductive setting. In our work, we experiment with both settings. Both of them demonstrate good performance.

Through above process, we can post-enhance all point cloud embeddings at inference time. In this way, the final retrieval stage is equivalent to adopting a re-ranking process. Through this re-ranking, the distribution of descriptors in the feature space would be more distinctive, hence the final place recognition results can be significantly increased.

\subsection{Loss}
To supervisely train or finetune the point cloud encoding network after the unsupervised pretraining, we adopt the following triplet loss to train our model on view of its superior performance in \cite{komorowski2021minkloc3d}:
\begin{equation}
L(v_a,v_p,f_n)=max\{d(v_a,v_p)-d(v_a,v_n)+m,0\}
\end{equation}
where $v_a$ is the descriptor of the query scan, $v_p$ and $v_n$ are descriptors of positive sample and negative sample respectively, and $m$ is a margin. $d(x,y)$ means the Euclidean distance between $x$ and $y$. To build informative triplets, we use batch-hard negative mining following \cite{komorowski2021minkloc3d}.

\begin{table*}[t]
\label{tab_baseline}
\centering
\begin{tabular}{l|cccc|cccc}
\hline
& \multicolumn{4}{c|}{{\bf Average recall at top-1\% (\%)}} &  \multicolumn{ 4}{c}{{\bf Average recall at top-1 (\%)}} \\
{\bf Method} & {\bf Oxford} & {\bf U.S.} & {\bf R.A.} & {\bf B.D.} & {\bf Oxford} & {\bf U.S.} & {\bf R.A.} & {\bf B.D.} \\
\hline
{\bf PointNetVLAD}  &       80.3 &       72.6 &       60.3 &       65.3 &          - &          - &          - &          - \\
{\bf PCAN}&       83.8 &       79.1 &       71.2 &       66.8 &          - &          - &          - &          - \\
{\bf DAGC} &       87.5 &       83.5 &       75.7 &       71.2 &          - &          - &          - &          - \\
{\bf SOE-Net} &       96.4 &       93.2 &       91.5 &       88.5 &          - &          - &          - &          - \\
{\bf SR-Net}  &       94.6 &       94.3 &       89.2 &       83.5 &       86.8 &       86.8 &       80.2 &       77.3 \\
{\bf LPD-Net}&       94.9 &         96.0 &       90.5 &       89.1 &       86.3 &         87.0 &       83.1 &       82.3 \\
{\bf Minkloc3D} &       97.9 &         95.0 &       91.2 &       88.5 &         93.0 &       86.7 &       80.4 &       81.5 \\
{\bf SVT-Net} &       97.8 & { 96.5} & { 92.7} & {90.7} &       93.7 & { 90.1} &  84.3 & 85.5 \\
\hline
{\bf GIDP+PointNet+Inductive} &      86.8 &	75.8&	72.1	&68.2&	73.6&	61.1&	57.8&	58.6\\
{\bf GIDP+PointNet+Transductive} &      88.7&	77.3&	73.3&	66.5&	77.1&	64.9&	60.8&	57.1\\
{\bf GIDP+DGCNN+Inductive} &     98.0&	98.1&	94.8&	91.5&	92.6&	 91.7&	87.8&	 \bf{86.0}\\
{\bf GIDP+DGCNN+Transductive} &     \bf{98.6}	& \bf{98.8}&	 \bf{95.5}&	 \bf{91.1}&	\bf{94.5}&	 \bf{93.5}	&\bf{90.5}&	85.3\\
\hline
\end{tabular}  
\caption{Comparison with state-of-the-art under the baseline setting.}
\end{table*}

\begin{table*}[ht]
\centering
\begin{tabular}{l|cccc|cccc}
\hline
& \multicolumn{4}{c|}{{\bf Average recall at top-1\% (\%)}} &  \multicolumn{4}{c}{{\bf Average recall at top-1 (\%)}} \\
{\bf Method}  & {\bf Oxford} & {\bf U.S.} & {\bf R.A.} & {\bf B.D.} & {\bf Oxford} & {\bf U.S.} & {\bf R.A.} & {\bf B.D.} \\
\hline
{\bf PointNetVLAD}  &       80.1 &       90.1 &       93.1 &       86.5 &       63.3 &       86.1 &       82.7 &       80.1 \\
{\bf PCAN}&       86.4 &       94.1 &       92.3 &         87.0 &       70.7 &       83.7 &       82.3 &       80.3 \\
{\bf DAGC}  &       87.8 &       94.3 &       93.4 &       88.5 &       71.4 &       86.3 &       82.8 &       81.3 \\
{\bf SOE-Net}  &       96.4 &       97.7 &       95.9 &       92.6 &       89.3 &       91.8 &       90.2 &         89.0 \\
{\bf SR-Net}  &       95.3 &       98.5 &       93.6 &       90.8 &       88.5 &       93.5 &       86.8 &       85.9 \\
{\bf LPD-Net} &       98.2 &       98.2 &       94.4 &       91.6 &         93.0 &       90.5 &       { 97.4} &       85.9 \\
{\bf Minkloc3D} &       98.5 &       99.7 &       99.3 &       96.7 & {94.8} &       97.2 &  96.7 &         94.0 \\

{\bf SVT-Net} &       98.4 & {\bf 99.9} & {\bf 99.5} &       \bf{97.2} &       94.7 &         97.0 &       95.2 &       \bf{94.4} \\
\hline
{\bf GIDP+DGCNN+Inductive} &     98.0&	99.7	&98.7&	96.2&	93.3&	96.4&	94.9	&93.1\\
{\bf GIDP+DGCNN+Transductive} &    \bf{98.6}&	99.8&	99.2&	95.8&	\bf{95.1}&	\bf{97.3}&	\bf{97.8}&	92.5\\
\hline
\end{tabular}  
\caption{Comparison with state-of-the-art under the refined setting.}
\label{tab_refine}
\end{table*}

\section{Experiments}
In this section, we first introduce the datasets we use. Then, we present the implementation details. Next, the comparison results are discussed. Finally, we present the ablation study.
\subsection{Dataset}
For fair comparison, we conduct experiments on the benchmark datasets proposed by \cite{uy2018pointnetvlad}, which is the most widely datasets for point cloud-based large-scale place recognition. Its benchmark consists of four datasets: one outdoor dataset named Oxford generated from Oxford RobotCar \cite{maddern20171} and three in-house datasets: university sector (U.S.), residential area (R.A.) and business district (B.D.). The four datasets contain 21711, 400, 320, 200 submaps for training and 3030, 80, 75, 200 submaps for testing for Oxford., U.S., R.A. and B.D. respectively. Each point cloud contains 4096 points. During training, point clouds are regarded as correct matches if they are at maximum 10m apart and wrong matches if they are at least 50m apart. In testing, the retrieved point cloud is regarded as a correct match if the distance is within 25m between the retrieved point cloud and the query scan. We use average recall at top 1\% and  average recall at top 1 as main metrics as previous methods \cite{uy2018pointnetvlad,fan2022svt}, for a fair comparison.

\subsection{Implementation Details}
We implement our method using PyTorch. We train two different version of models using PointNet and DGCNN as backbones respectively.  At pretraining state, the batch size is 64. The learning rate is 0.03.  The model is pretrained for 100 epochs using Adam optimizer. The feature dimension of both $u_i$ and $v_i$ is 256. At encoding network supervised training stage, we train our model under two setting: the baseline setting which only uses training set of Oxford to train the model, and the refined setting which additionally add training set of U.S. and R.A. In the baseline setting, the initial batch size is 32 and the initial learning rate is $10^{-3}$. The model is trained for 40 epochs and the learning rate is decayed by 10 at the end of the 30th epoch.  The refined model is trained with an  initial batch size of 16 and an initial learning rate of $10^{-3}$. The model is trained for 80 epochs and the learning rate is decayed by 10 at the end of the 60th epoch. At the final re-ranking based post-enhancing stage,  $\lambda$ is set to 0.2. The dimension of the final global point cloud descriptor is 256. The number of neighbors $K$ is 5. All experiments are conducted on a single A6000 GPU.

\subsection{Results}
In this section, we conduct experiments on Oxford., U.S., R.A. and B.D. and compare our method with the state-of-the-art methods include PointNetVLAD \cite{uy2018pointnetvlad}, PCAN \cite{zhang2019pcan}, DAGC \cite{sun2020dagc}, SR-Net \cite{fan2020srnet}, LPD-Net \cite{liu2019lpd}, SOE-Net \cite{xia2021soe}, Minkloc3D \cite{komorowski2021minkloc3d} and SVT-Ne \cite{fan2022svt}. PointNet and DGCNN are used are backbones respectively. We show results on both baseline setting and refined setting. We also show results on both Inductive setting and Transductive setting.

\noindent \textbf{Results on the baseline setting:} In Table I, we show the results on the baseline setting. We can find from the table that  1) Even using the simplest baseline PointNet, our method performs very well. PointNetVLAD also use PointNet as their backbone network, while our method outperforms it for a large margin, showing the superiority of our method. Our method using PointNet even outperform DAGC that uses DGCNN as the backbone, greatly demonstrating the effectiveness of our method. 2) When using DGCNN as the backbone, our method outperforms all previous methods and achieves new stae-of-the-art. Note DGCNN is very simple and light-weight backbone compared backbone used by other methods. 3) The performance of our method under the Transductive setting outperforms that of the Inductive setting. That is because under the  Transductive setting, other point clouds collected at the inference time are used for descriptor post-enhancing, bringing more semantic information and geometry information to the query descriptor. 4) Stronger backbones can bring better performance. It can been seen that the model using DGCNN performs significantly better than the model using PointNet. That is because DGCNN is stronger in learning local features, therefore it will benefit more from our pretraining stage and post-enhancing stage. In a word, our GIDP is superior than all existing methods though only simple backbones are used. We contribute the superiority to the proposed unsupervised  momentum contrast point cloud pretraining module and the re-ranking based descriptor post-enhancing module.

\noindent \textbf{Results on the refined setting:} In Table \ref{tab_refine}, we show the refined results of our method and other methods. In this experiment, we only show the result of using DGCNN as backbone. It can be found that after adding the training set of indoor datasets U.S. and R.A., the performance of all methods are improved. Under the Transductive setting, our method performs comparable with the current state-of-the-art model SVT-Net, though we use a much simpler backbone while SVT-Net use sparse Transformers. Under the Transductive setting, benefited from the additional inference time captured point clouds, our method wins 3 out 4 datasets at the more strict average recall at top 1 metric. This demonstrates the superiority of our re-ranking based post-enhancing module. However, we also find that the difference between our method and SVT-Net is tiny. This is partly because the current dataset is too small so the performance become saturated. Therefore, in the future work, a new more large-scale dataset should be built to benchmark novel methods.

\begin{table}
\begin{tabular}{l|cccc}
\hline
           & \multicolumn{ 4}{|c}{{\bf Avg recall at top 1\%}} \\

     Variants     & {\bf Oxford} & {\bf U.S.} & {\bf R.A.} & {\bf B.D.} \\
\hline
{\bf Random Init+PoinNet} &       83.0 &       69.5 &       72.5 &       62.5 \\

{\bf Random Init+PoinNet+Inductive} &       87.3 &       70.9 &       74.7 &       62.0 \\

{\bf Random Init +PoinNet+Transductive} &       85.1 &       69.1 &       72.5 &       63.2 \\

{\bf GIDP+PoinNet+Inductive} &       86.8 &       75.8 &       72.1 & {\bf 68.2} \\

{\bf GIDP+PoinNet+Transductive} & {\bf 88.7} & {\bf 77.3} & {\bf 73.3} &       66.5 \\
\hline
\end{tabular}  
\caption{Results of ablation study}
\label{tab_ablation}
\end{table}

\begin{table}
\begin{tabular}{l|cccc}
\hline
    {\bf } & \multicolumn{ 4}{|c}{{\bf Avg recall at top 1\%}} \\

{\bf variants} & {\bf Oxford} & {\bf U.S.} & {\bf R.A.} & {\bf B.D.} \\
\hline
{\bf without Jitter} &       98.0 &       98.1 &       93.0 &       91.3 \\

{\bf without Randompoints Removal } &       97.5 &       97.2 &       93.1 &       89.1 \\

{\bf without Randomblock Removel} &       97.8 &       97.0 &       94.1 &       90.3 \\

{\bf without RandomShear} &       97.6 &       97.2 &       92.7 &       88.8 \\

{\bf GIDP+DGCNN+Iransductive} & {\bf 98.6} & {\bf 98.8} & {\bf 95.5} & {\bf 91.1} \\
\hline
\end{tabular}  
\caption{Impact of data augmentation in unsupervised pretraining.}
\label{tab_aug}
\end{table}

\begin{table}
\begin{tabular}{l|cccc}
\hline
           & \multicolumn{ 4}{|c}{{\bf Avg recall at top 1\%}} \\

      variants     & {\bf Oxford} & {\bf U.S.} & {\bf R.A.} & {\bf B.D.} \\
\hline
without Projection Head &       97.6 &       96.2 &       94.6 &       89.5 \\

{\bf GIDP+DGCNN+Iransductive} & {\bf 98.6} & {\bf 98.8} & {\bf 95.5} & {\bf 91.1} \\
\hline
\end{tabular}  
\caption{Impact of projection head in unsupervised pretraining.}
\label{tab_projection}
\end{table}

\begin{table}
\begin{tabular}{l|cccc}
\hline
           & \multicolumn{ 4}{|c}{{\bf Avg recall at top 1\%}} \\

   variants        & {\bf Oxford} & {\bf U.S.} & {\bf R.A.} & {\bf B.D.} \\
\hline
{\bf GIDP+PoinNet+Inductive} &       86.8 &       75.8 &       72.1 &       68.2 \\

{\bf GIDP+DGCNN8+Inductive} &       97.6 &       96.2 &       92.8 &       87.8 \\

{\bf GIDP+DGCNN16+Inductive} & {\bf 97.6} &       96.2 & {\bf 94.6} &       89.5 \\

{\bf GIDP+DGCNN24+Inductive} &       97.4 & {\bf 97.0} &       93.8 & {\bf 90.3} \\
\hline
\end{tabular}  
\caption{Impact of different backbones}
\label{tab_backbone}
\end{table}

\subsection{Ablation study}
In this section, we study the impact of our proposed models and other key designs. All experiments are conducted at the baseline training setting.

\noindent \textbf{Effectiveness of unsupervised  momentum contrast point cloud pretraining module:} The unsupervised  momentum contrast point cloud pretraining module is one of the key contributions of our work. This module plays a role of learning a good initialization for the point cloud encoding network. In Table \ref{tab_ablation}, we show the result of without using this module. It can be found that without this module, the performance of the model is decreased significantly. This greatly demonstrate the importance of introducing a pretraining stage for large-scale place recognition.

In the unsupervised  momentum contrast point cloud pretraining module, we use random data augmentation to generate positive samples. In Table \ref{tab_aug}, we show the results of removing one kind of data augmentation each time. It can be found from the table that all four kinds of random data augmentation plays a significant role of improving performance. It is because these data augmentation methods can simulate different situations of capturing the same scene with different geometries using LiDAR,  so the model can learn more general features through pretraing.

In the pretraining stage, as introduced before, we  adopt a projection head after the point cloud encoding network. In Table \ref{tab_projection}, we investigate the impact of the projection head. It can be seen that without the projection head. The performance decreases a lot. This demonstrates that the effectiveness of adding the projection head, which increases the generalization ability of the pretrained weights.

\noindent \textbf{Effectiveness of the re-ranking based descriptor post-enhancing module:} In the re-ranking based descriptor post-enhancing module, each descriptor is interact with its neighbors in the feature space to enhance its representational ability. In the top 3 rows of Table \ref{tab_ablation}, we show the impact of this module. To avoid the influence of the unsupervised  momentum contrast point cloud pretraining module, we randomly initialize  point cloud encoding network during training. PointNet is used as the backbone. It can be seen from the table that without this module, the model performs relatively poor. After adding  the module, the performance of the model increases for a large margin. Then, if we use the Transductive setting, the result is further increased. These experiments greatly demonstrate the correctness of designing the  re-ranking based descriptor post-enhancing module.

\noindent \textbf{Impact of different backbones:} In Table  \ref{tab_backbone}, we show the performance of our method with different backbones. It can be seen that stronger backbones can bring better results. For the DGCNN backbone, we test the results of indexing different number of neighbors when learning local features. It can be found that with the increasing of number of neighbors, the performance increases. However, when it is larger than 16, the improvement is tiny. Therefore, we choose 16 as our default setting.

\section{CONCLUSIONS AND LIMITATION}
In this paper, we propose a novel method named GIDP to learn a good initialization and inducing descriptor post-enhancing for point cloud based large-scale place recognition. In GIDP, an unsupervised momentum contrast point cloud pretraining module and a re-ranking based descriptor post-enhancing module are proposed to achieve our goal. We have conducted extensive experiments on both indoor and outdoor datasets. Results show that our method can achieve state-of-the-art performance using simple backbones.

Though our method achieves good performance, there still exist limitations. For example, currently, we only investigate pretrain point-based backbones, how to pretrain other type of backbones, such as sparse voxel-based backbones, is not studied. In the future, we plan to investigate more novel pretrain methods and make the training pipeline more efficient.

\addtolength{\textheight}{-3cm}   




\bibliographystyle{IEEEtran}
\bibliography{IEEEfull}

\end{document}